\documentclass{fcs}
\usepackage{bm}
\usepackage{graphicx}
\usepackage{caption}
\usepackage{subfigure}
\usepackage{booktabs}
\graphicspath{{figures/}}

\volumn{ }
\doi{ }
\articletype{RESEARCH~ARTICLE}
\copynote{{\copyright} Higher Education Press and Springer-Verlag Berlin Heidelberg 2012}
\ratime{Received month dd, yyyy; accepted month dd, yyyy}
\email{guangcunshan@hotmail.com\\ $\dagger$~Tian Wang and Shiye Lei have the same contribution to this work.}
\title{Accelerating~temporal~action~proposal~generation~via \\ high~performance~computing}
\author{Tian Wang $^{1,\dagger}$, Shiye Lei$^{1,\dagger}$, Youyou Jiang$^{2}$, Choi Chang$^{3}$, Hichem Snoussi$^{4}$, Guangcun Shan \xff $^{5}$}
\address{{1\quad School of Automation Science and Electrical Engineering, Beihang University, Beijing 100191, China}\\
{2\quad  School of Software, Tsinghua University, Beijing 100084, China}\\
{3\quad  Department of Computer Engineering, Gachon University, Seongnam 1342, Korea}\\
{4\quad Institute Charles Delaunay-LM2S-UMR STMR 6281 CNRS, University of Technology of Troyes, Troyes 10010, France}\\
{5\quad  School of Instrumentation Science and Opto-electronics Engineering, Beihang University, Beijing 100191, China}}
\begin{document}
\maketitle
\setcounter{page}{1}
\setlength{\baselineskip}{14pt}


\begin{abstract}
Temporal action recognition always depends on temporal action proposal generation to hypothesize actions and algorithms usually need to process very long video sequences and output the starting and ending times of each potential action in each video suffering from high computation cost. To address this, based on boundary sensitive network we propose a new temporal convolution network called Multipath Temporal ConvNet (MTN), which consists of two parts i.e. Multipath DenseNet and SE-ConvNet. In this work, one novel high performance ring parallel architecture based on Message Passing Interface (MPI) is further introduced into temporal action proposal generation, which is a reliable communication protocol, in order to respond to the requirements of large memory occupation and a large number of videos. Remarkably, the total data transmission is reduced by adding a connection between multiple computing load in the newly developed architecture. It is found that, compared to the traditional Parameter Server architecture, our parallel architecture has higher efficiency on temporal action detection task with multiple GPUs, which is suitable for dealing with the tasks of temporal action proposal generation, especially for large datasets of millions of videos. We conduct experiments on ActivityNet-1.3 and THUMOS14, where our method outperforms other state-of-art temporal action detection methods with high recall and high temporal precision. In addition, a time metric is further proposed here to evaluate the speed performance in the distributed training process.
\end{abstract}

\Keywords{temporal convolution, temporal action proposal generation, deep learning.}

\section{Introduction}
With the rapid development of the Internet and camera, the number of videos is increasing at a very high speed. There are millions of video submissions on video-sharing websites like YouTube every day. Besides, the video surveillance system plays an important role in maintaining security \cite{8253857}\cite{8733112}\cite{wang2018generative}. These video files contain a lot of information for human, such as time duration and action classify \cite{8709763}\cite{wang2019aed}. Making full use of videos is an indispensable step for building a smart city. It is vital for the development of information age to extract information from a large number of videos by automatically. Action is the most important information for videos because the essence of the video is recording varieties of motion. So a significant branch of video task is action recognition, which aims to recognize the class of action from a trimmed video. But the task is limited because its research object is videos that have been manually trimmed and only contain single action. The majority of videos in the real world are untrimmed videos and contains multiple action instances in a single video. The problem requires another challenging task: temporal action detection, which aims to recognize the temporal boundaries and classes of action instance from untrimmed videos.

Temporal action detection usually includes two steps: proposal and classification. Proposal stage focuses on detecting action boundary and generating action instance with untrimmed video. Classification is aim to recognize the class of action instance produced in the previous step. For the task of temporal action detection, classification has achieved high accuracy. And the precision of proposals is the main factor limiting temporal action detection \cite{caba2015activitynet}\cite{jiang2014thumos}.

High-quality proposals should meet two requirements \cite{lin2018bsn}: (1) high recall; (2) high overlap with ground truth. And a good algorithm of generating proposal should not only generate excellent proposals, but its speed should be as fast as possible. Because videos occupy a large amount of memory,  and we must improve the speed of method so for being applied to practice. 

Most proposal generation algorithms generate generation by using sliding windows \cite{buch2017sst}\cite{caba2016fast}\cite{escorcia2016daps}\cite{shou2016temporal}. But the pre-defined durations and intervals of sliding windows cannot generate proposals with flexible length, which greatly reduced the precision of proposals. Boundary Sensitive Network (BSN) \cite{lin2018bsn} used a temporal network with 3 convolution layers to deal with video feature sequences and could generate proposals with flexible duration. But BSN cannot extract enough information from videos due to its simple network architecture.

To address these problems and improve the quality of producing proposals, we designed a temporal convolution network architecture, which adopted two channel convolution for extracting both temporal and spatial information from video feature sequence. While we made the network architecture more complicated, a new parallel computing framework was used to accelerate our algorithm with higher efficiency compared to the popular Parameter-Server Framework \cite{dean2012large}.

In summary, the main contribution of our work is three-fold:
\begin{enumerate}
\renewcommand{\labelenumi}{(\theenumi)}
\item We proposed Multipath Temporal Network that could extract effective information from video feature sequence.
\item We adopted a new parallel computing framework to speed up our temporal convolution network with high efficiency.
\item A metric is put forward to evaluate the time consumption in the distributed deep learning field.

\end{enumerate}

\section{Related work}
Temporal action detection aims to detect action instance from the untrimmed video. The task could be divided into two steps: proposal and classification. Though some methods do the two steps at the same time, the majority of methods take the task as a serial process and finish proposal and classification separately.

\textbf{Temporal action proposal generation.} Proposal generation is the distinct characteristic of Temporal action recognition. Proposal generation aims to detect the start and end boundary of action instance in the untrimmed video. Earlier methods used sliding windows to generate proposals \cite{karaman2014fast}\cite{wang2014action}. Then some algorithms \cite{buch2017sst}\cite{gao2017turn}\cite{caba2016fast}\cite{escorcia2016daps} began to pre-define temporal duration and intervals of proposals, and evaluated them with multiple methods like recurrent neural network (RNN) and dictionary learning. Another popular method for proposal generation is TAG \cite{zhao2017temporal}, which utilized watershed algorithm to do the project. Though TAG can generate proposals with flexible boundaries and durations, it is lack of evaluation to these proposals. BSN \cite{lin2018bsn} has a good performance of generating proposals, which is benefit from its temporal convolution network. But the weak extraction capacity to video feature sequence because of the single temporal convolution network and slow speed make it difficult to be applied to practice.     

\textbf{Action recognition.} The classification of action instances is as the same as the task of action recognition. Before the wide range applied of deep learning in computer vision, improved Dense Trajectory (iDT) \cite{wang2011action} has a very good performance in action recognition. It adopted manual image features such as Histogram of Oriented Optical Flow (HOF), Histogram of Oriented Gradient (HOG) and Motion Boundary Histograms (MBH), and used Fisher Vector to encode these features. Then an SVM classifier was designed to classify the features encoded. With the rapid development of deep learning, convolution neural network (CNN) brought great effects to computer vision and showed strong strength in action recognition. Two-stream network \cite{feichtenhofer2016convolutional} has two parts and extracts appearance features from RGB frame with using spacial CNN and extracts motion features from optical flow field with using temporal ConvNet. TSN \cite{wang2016temporal} improved two-stream network by using multiple networks to capture short-term temporal information. C3D network \cite{tran2015learning} is different from two-stream network. It is 3-dimensional and extract features from raw videos directly. There are also lots of 3D convolution structure be proposed for extracting more information from videos.

\textbf{Distributed deep learning} Because deep learning has a wide range of application, acceleration is significant to let it more widely to be used. Distributed deep learning, which is based on parallel computing, belongs to high performance computing and accelerate CNN by using more machines like GPUs. 

MapReduce\cite{dean2008mapreduce} was proposed by Google and dissemble compute into map and reduce, which divided compute into tow steps of Map and Reduce. But it has a strict requirement of consistency. To address the problem, Graphlab\cite{low2012distributed} used an abstract way like the image to communicate, which also lead to low scalability. Jeff Dean proposed Parameter-Server Framework (PS) \cite{dean2012large}, which uses a parameter server to store the newest weight parameters of CNN. When the number of GPUs increases, the efficient of PS will have a great decline because of the big communication.

Video task is closed to real life and Allied in practice is its final goal. So besides accuracy, speed of methods is also an important indicator. Based on these, our method is superior to others in two aspects: (1) Our improved temporal convolution has a more reasonable architecture for video files and could extract more useful information from video feature sequence; (2) We combine our proposal generation method with a new framework of parallel computing for efficient acceleration.

\section{Proposal generation}
Based on SENet (Squeeze-and-Excitation Network) \cite{hu2018squeeze} and exploring the meaning of video feature sequence, we proposed Multipath Temporal Network (MTN).
\begin{figure*}[ht]
    \centering
    \includegraphics[width=0.9\textwidth]{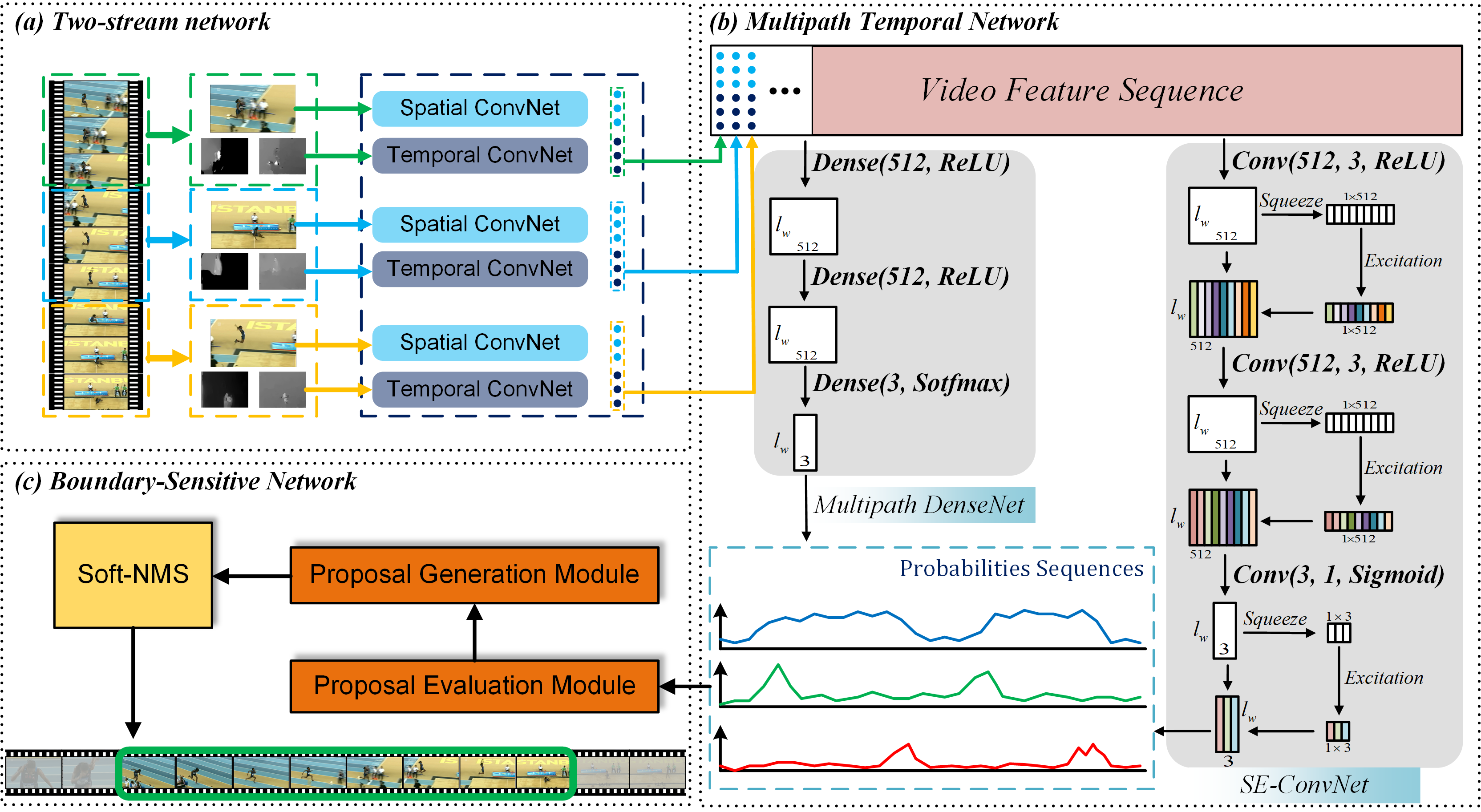}
    \caption{The framework of our approach. (a) Two-stream network is used to encode visual features in snippet-level. (b) The architecture of Multipath Temporal Network:\textit{SE-ConvNet} extract temporal information from video feature sequence; \textit{Multipath DenseNet} uses multiple dense layers to explore the meaning of single feature vector. (c) Boundary-Sensitive Network is used for generating proposals by probabilities sequence}
    \label{mtn}
\end{figure*}

\subsection{Video feature sequence}
For lots of tasks in video analysis, they do not handle video directly but deal with video feature sequence. Video feature sequence is usually encoded by neural network with special structure. In this paper, we used two-stream network \cite{simonyan2014two} as an encoder to transfer video into a set of vectors. 

In detail, the architecture of two-stream network is shown as Fig. \ref{mtn} (a). The network contains two part: spacial network extracts information from RGB images and temporal network is used to draw importance from optical flow images. 

We selected 1 RGB image and 2 optical images from every 16 frames and put them into two-stream network. The outputs of spacial network and temporal network are both 200-dimensional vector. Through concatenating them, we can get a 400-dimensional vector for the 16 frames finally. For a video with $n$ frames, two-stream network produced $n/16$ 400-dimensional vectors, which are video feature sequence.

We denote an untrimmed video with $N$ frames as $X=\{x_n\}_{n=1}^N$. For $X$, then two-stream network produced $N/16$ 400-dimensional vectors represented as $V=\{v_n\}_{n=1}^{N/16}$.

To normalize our inputs, Interpolation is applied to convert the number of vectors, $N/16$, to $100$ in our experience. So far, a untrimmed video was transformed into $V=\{v_n\}_{n=1}^{100}$, and $V$ is the input of our temporal convolution network.

\subsection{Temporal convolution network}
\begin{figure*}[ht]     
    \centering
    \includegraphics[width=0.9\textwidth]{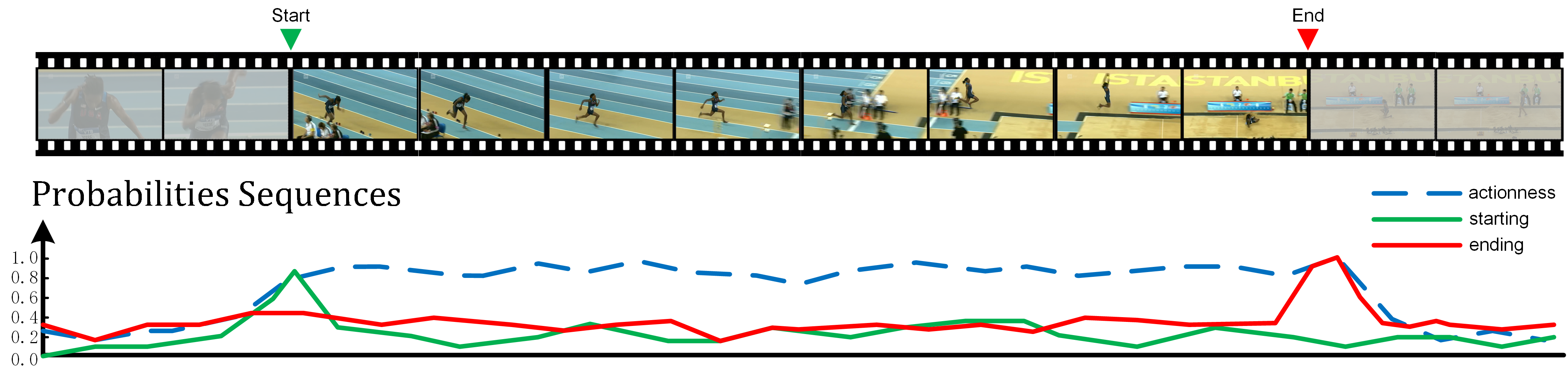}
    \caption{Temporal convolution network generate probabilities sequences}
    \label{probabilities_sequence}
\end{figure*}

To generate flexible proposals, a temporal convolution network is usually used to detect possibility with each $v_n$ in $V$, like Fig. \ref{probabilities_sequence}. For a $V=\{v_n\}_{n=1}^{100}$, temporal convolution network generates 3 possibility sequence $P_s=\{p_n^s\}_{n=1}^{100}, P_a=\{p_n^a\}_{n=1}^{100}, P_e=\{p_n^e\}_{n=1}^{100}$. $p_n^s, p_n^a, p_n^e$ present the possibility of action start, actionness and end in the duration of $v_n$ respectively.

The traditional temporal convolution network uses 3 one-dimensional convolution network to extract information from video feature vectors. A video feature sequence can be seen as a vector which length is $100$ and the channel is $400$. A temporal convolution layer can be denoted as $Conv(n_f, n_k, act)$, where $n_f, n_k, act$ denote the number of filters, kernel size and activation function, respectively. So traditional temporal convolution network could be defined as $Conv(512, 3, ReLU) \rightarrow Conv(512, 3, ReLU) \rightarrow Conv(3, 1, Sigmoid)$. 

The network can acquire temporal information from feature sequence, but its simple network is not enough power to extract lots of complicated information in the video. It just focuses on the temporal information of video feature sequence but overlooks the meaning of the single feature vector in video feature sequence.


Inspired by the ability to extracting information from multiple channel feature maps of SENet and excellent dimensional representation of dense connect layer, we proposed Multipath Temporal Network (MTN), which could better extract information from video feature sequence. Fig. \ref{mtn} (b) shows the structure of MTN. There are two networks of Multipath DenseNet and SE-ConvNet in multipath temporal network. Multipath DenseNet with 3 dense connect layers is used to detect the deep meaning of the single feature vector one by one. 

In the original convolution process, lots of feature maps are produced by using a large number of convolution kernel. Beside, because of one-dimension, only a few information is contained in a single feature map, which increases the reliance of network on raw input data. In convolution layer, all of feature maps in the same layer have the same weights for the next layer. But some of feature maps contain more effective information compared with others in the same convolution layer. If we could pay more attention to these effective feature maps, and ignore useless feature maps properly, our temporal convolution network will have a better effect and its robustness will be more strong.

we improved the ConvNet by using the squeeze-excitation block. In squeeze, global pooling is applied to compress feature maps on the spatial dimension. Each feature map will be transformed into a single number and the number of channels is constant. Then we put the output of squeeze into dense connected neural network and generate weights for all the feature maps, which is called excitation. The shape of output and input in excitation stage are consistent. Multiply weights for these feature maps so we get the final result of ConvNet. We apply the squeeze-excitation block after every convolution layer in our temporal convolution architecture to enhance the ability to extracting information.

While squeeze-excitation blocks are used to enhance our one-dimension convolution layer, we only extract temporal information between video feature sequences. For a single video feature vector, we need to know what meaning it represents. In order to achieve this goal, we added multiple dense layers named Multipath DenseNet. By using it, we can extract information from video feature sequence on the spatial dimension. For our Multiple DenseNet, the number of units in input layer is $400$, corresponds to the dimension of feature vectors. The number of unit in hidden layer is $512$ with $ReLU$ as the activation function. The number of unit in output layer is $3$ with $softmax$ activation function for outputting the probability of action start, end, and actionness. Dense layer could be presented as $Dense(units, Act)$, where $units$ and $Act$ are the number of units and activation function of dense connected layer. So Multipath DenseNet can be defined as $Dense(512, ReLU) \rightarrow Dense(512, ReLU) \rightarrow Dense(3, Softmax)$. 

We can see that our improvement for temporal convolution network is not in depth but width. SE-ConvNet and Multipath DenseNet deal with video feature sequence separately and get their own probabilities sequence. Then through adding them according to a weight parameter $W$, we can get the final probabilities sequences $P_S, P_E, P_A$.

\subsection{Training of temporal convolution network}
Because the output of temporal convolution network is 3 probabilities sequences, the overall loss function consists of three parts, which is as below:
\begin{equation}
    Loss = \lambda \cdot L_{action} + L_{start} + L_{end}
\label{loss_function}
\end{equation}
where $\lambda$ is a weight and set to $2$ in our network.

To compute Loss, we need to convert the ground truth to the label for training our network. The duration between $v_n$ and $v_{n+1}$ is denoted as $l_w$, which is equal to $l/100$, where $l$ is the length of video. The moment of $v_n$ is $t_n$, so we define the region of $v_n$ is $r_n=[t_n-l_w/2, t_n+l_w/2]$. And now we get the set of region $R=\{r_n\}_{n=1}^N$. Besides, There is only a instant moment for the start and end of a action, for example, $t_s$ and $t_e$, we also need to transfer them into regions $r_s=[t_s-l_w/2, t_s+l_w/2]$ and $r_e=[t_e-l_w/2, t_e+l_w/2]$. For each region $r_n$, it consists of three parts: start, actionness and end. Then we can get the label $g_n=(g_n^s, g_n^a, g_n^e)$ for each $v_n$, where $g_n^s$ and $g_n^e$ are the proportion of $r_s$ and $r_e$ in $r_n$ and $g_n^s=1-g_n^s-g_n^e$. 

We adopt cross entropy to compute our loss function:
\begin{equation}
L=\frac{1}{l_{w}} \sum_{i=1}^{l_{w}}\left(\alpha^{+} \cdot b_{i} \cdot \log \left(p_{i}\right)+\alpha^{-} \cdot\left(1-b_{i}\right) \cdot \log \left(1-p_{i}\right)\right)
\end{equation}
where $b_i=sign(g_i-\theta_{IoP})$ is a two-values function and $\theta_{IoP}$ is set to 0.5 in our network. $l^+=\sum g_i$ and $l^-=l_w-l^+$. we also introduced $\alpha^{+}=\frac{l_{w}}{l-}$ and $\alpha^{-}=\frac{l_{w}}{l^{+}}$ to balance the error caused by the imbalance between the number of positive and negative samples.

\section{Parallel computing acceleration}
\subsection{Classes of parallel computing}
\begin{figure}[ht]
    \centering
    \includegraphics[width=0.45\textwidth]{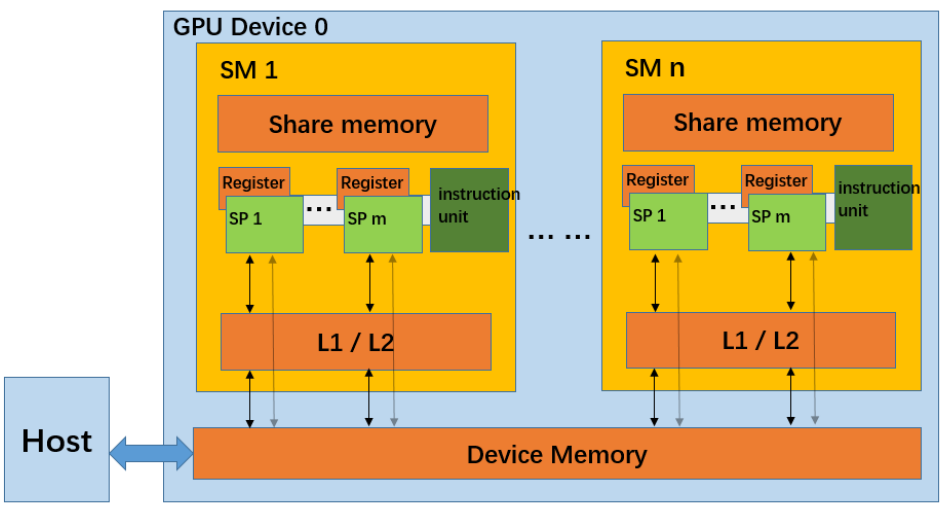}
    \caption{Hard architecture of GPU}
    \label{gpu-architecture}
\end{figure}

For training of the neural network, GPU is a much faster platform than CPU because of its architectural advantage on matrix computation. So we used GPU as the computing platform for our temporal network. Further, we applied parallel computing based on GPU to accelerate our network for more powerful capabilities of processing video. The hard architecture of GPU is shown as Fig. \ref{gpu-architecture}. Stream Processor (SP) is the basic computing unit of GPU and Stream Multiprocessor (SM) is composed of a certain number of SM, register, share memory and L1/L2 cache. From the picture, we can see that one SM contains multiple SP but only one instruction unit. So for single SM, it only supports single instruction multiple data (SIMD) but not multiple instruction multiple data (MIMD). When GPUs are used to train CNN models, parameters of models are stored at device memory, which decided that the form of parallel computing of GPU in deep learning field is SIMD. After determining SIMD, the parallel architecture of GPU can be divided into two classes of model parallelism and data parallelism.

\subsection{Model parallelism}
\begin{figure}[ht]
    \centering
    \includegraphics[width=0.3\textwidth]{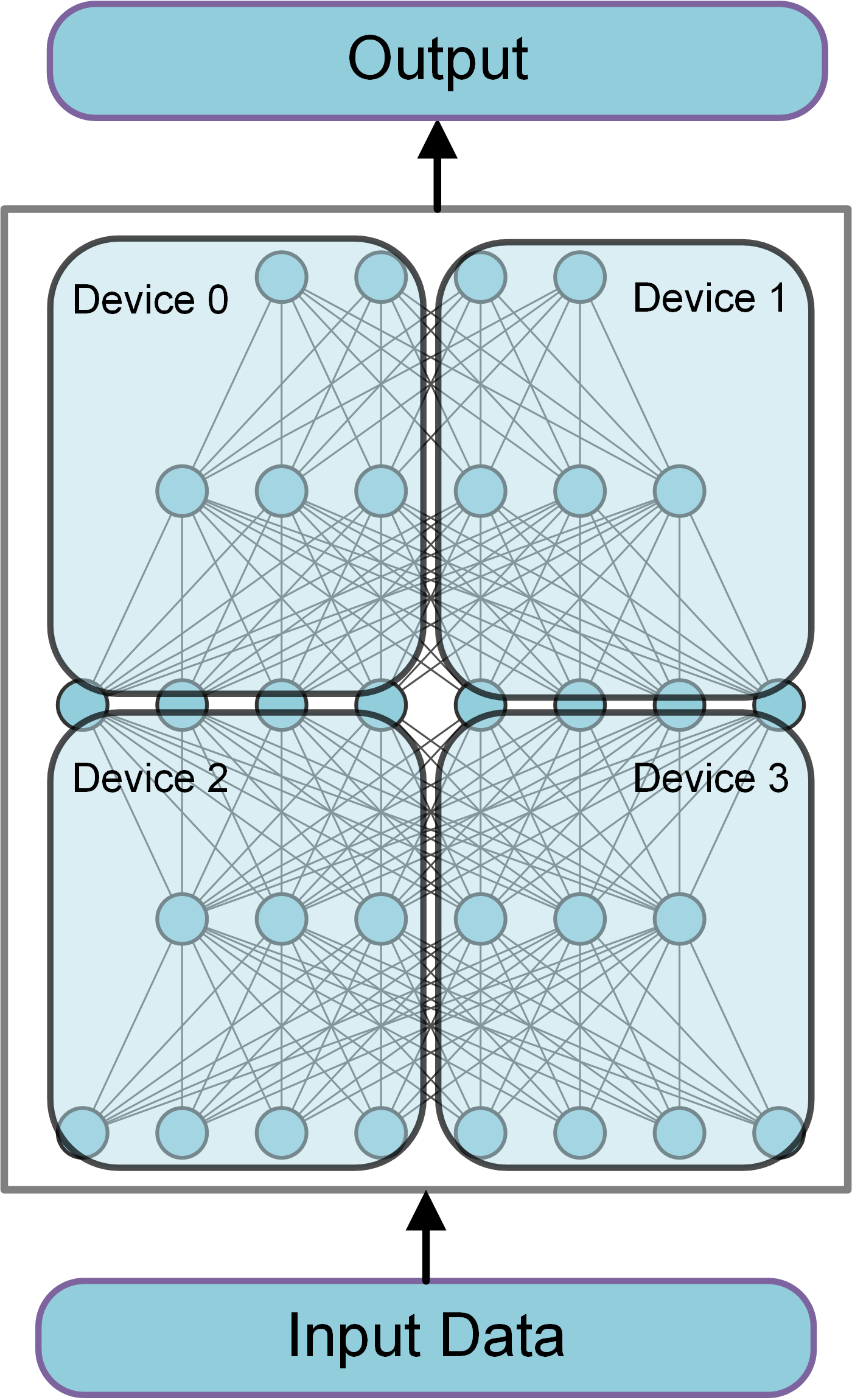}
    \caption{Model parallelism}
    \label{model_parallelism}
\end{figure}

Model parallelism means different machines (GPU or CPU) in a distributed system are responsible for different parts of a network model. For example, different layers in a neural network model or different parameters in the same layer are assigned to different machines. The structure of model parallelism is shown as Fig. \ref{model_parallelism}.
In general, the reason for applying model parallelism is oversized for the neural network.

\subsection{Data parallelism}
Data parallelism means the input data are divided into several parts and delivered to different machines. There is a complete model in each machine and these machines run the same program to deal with allocated data. Training CNN is a serial process, i.e., only after computing the gradients for current data and upgrading parameter weights, the next data can be put into the machine. The key of model parallelism is that all of GPUs have the same CNN model, which we called them model replicas. But the data for each GPU is different. We integrate different weight gradients $\nabla w$ calculated by all of GPUs and upgrade parameters of the model.

\begin{figure}[ht]
    \centering
    \includegraphics[width=0.45\textwidth]{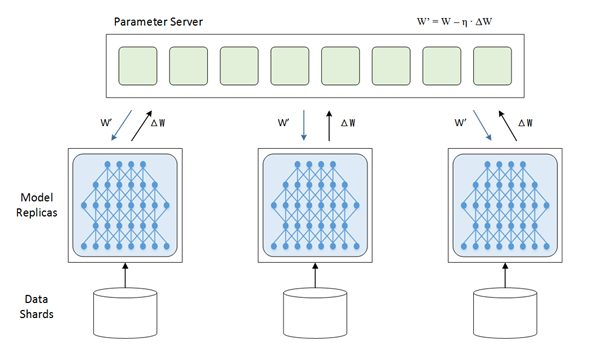}
    \caption{Parameter Server framework}
    \label{data-parallelism}
\end{figure}

With its simple and understandable structure, Parameter Server (PS) becomes the main data parallelism framework and got support from some mainstream deep learning framework like TensorFlow \cite{abadi2016tensorflow}. The architecture of PS is shown as Fig. \ref{data-parallelism}, where $\Delta w$ is the weight gradients computed by model replicas like GPUs and $w'$ is the newest weight parameters. PS stores the parameters of the model. Model replicas compute different parameter weights and then upgrade the parameters in the parameter server.

\subsection{Ring parallel architecture}
From Fig. \ref{data-parallelism} we can see that the communications volumes increase linearly with the increasing number of GPUs. We suppose the size of CNN model is $M$ and $N$ GPUs are used in our distributed system, so the communications volume is $N\cdot M$. If the number of GPUs achieved a high level, the large communications volume will greatly limit the training speed of CNN model.

\begin{figure}[ht]
    \centering
    \includegraphics[width=0.35\textwidth]{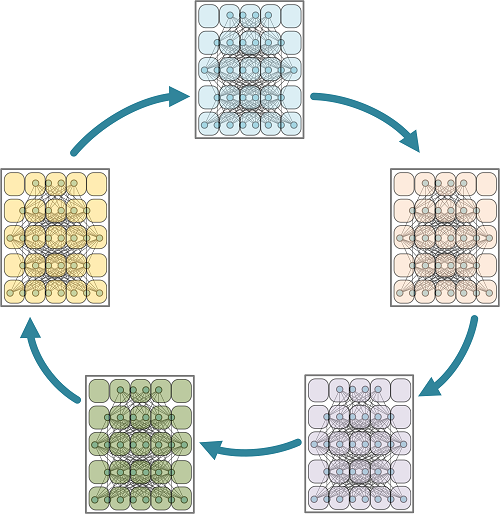}
    \caption{Ring parallel architecture. Multiple colors denote the weight gradient computed by each GPU is different.}
    \label{Ring-architecture}
\end{figure}

To address the problem of large communication volume in distributed deep learning system, we proposed ring parallel architecture. By building communication between GPUs with Message Passing Interface (MPI), our ring parallel architecture can reduce the pressure of communication. The ring parallel architecture is shown as Fig. \ref{Ring-architecture}. We changed the parallel topology to ring and averaged the communications volumes. The ring architecture upgrades weights through two steps including scatter and gather. 
 
\subsubsection{Scatter}
\begin{figure}[ht]
    \centering
    \includegraphics[width=0.35\textwidth]{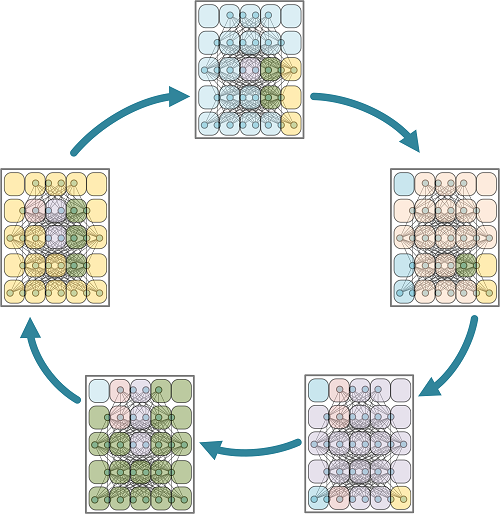}
    \caption{Scatter. In the scatter step, the GPU passes a row of weight gradient (all of the colors in this row) to the same position in its next GPU.}
    \label{ring-process}
\end{figure}

We divide weights in every GPU into $N$ parts, where $N$ is the number of GPU utilized in the architecture. After all of GPUs got different weight gradients by computing different input data, like there are different colors in Fig. \ref{Ring-architecture} and a row of colored blocks denotes a part. The $n$-th GPU passes its own $(n-i)\%N$-th block of weight gradients to its right neighbor and receives $(n-i-1)\%N$-th block of weight gradients from its left neighbor, where $i$ is the round of scatter. Fig. \ref{ring-process} shows the detail after one round of scatter.

\begin{figure}[ht]
    \centering
    \includegraphics[width=0.35\textwidth]{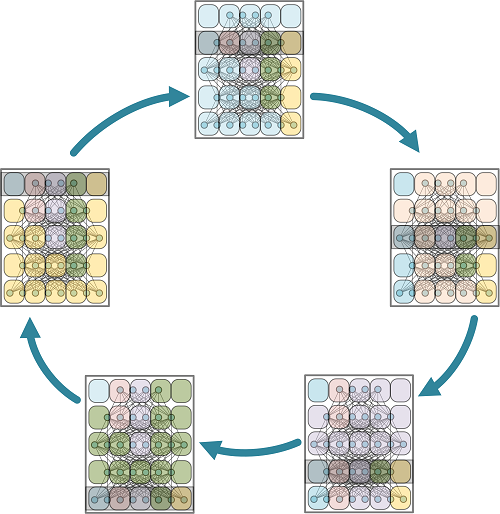}
    \caption{Distribution of weight gradients after scatter. There is a row in every GPU that has collected weight gradients from all of GPUs (all of the colors).}
    \label{scatter-reduce}
\end{figure}

After $N-1$ rounds of scatter, $n$-th GPU has collected $(n+1)\%N$-th block of weight gradients from all GPUs, which is shown as Fig. \ref{scatter-reduce}. After scatter, each GPU has a block of gradients which is from all GPUs.

\subsubsection{Gather}
\begin{figure}[ht]
    \centering
    \includegraphics[width=0.35\textwidth]{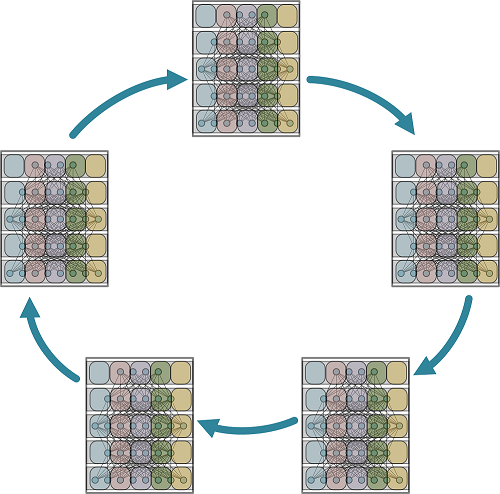}
    \caption{Distribution of weight gradients after gather. All of the weight gradients are merged in each GPU.}
    \label{all-gather}
\end{figure}

Like scatter, GPUs also pass a block of weight gradients to the next GPU in the process of gather. Through $N-1$ rounds of gather, the $(n+1)\%N$-th block of weight gradients in the $n$-th GPU is passed to all of other GPUs. In the $i$-th round of gather, the $n$-th GPU passes its own $(n-i-1)\%N$-th blocks of weight gradients to its right neighbor and receives the $(n-i-2)\%N$-th blocks of weight gradients from its left neighbor. Different from scatter, GPUs don't need to add but replace its own block by the block received. After gather, we can see that all of GPUs have obtained all weight gradients computed by every GPU, which is shown as Fig. \ref{all-gather}.

\subsection{Training time metrics}
To explore the relationship between the number of GPUs and training time and evaluate parallel architecture, we defined training time metrics $T(n)$, where $n$ is the number of GPU used.

Training time in distributed deep learning system could consist of three parts:(1) $t_1$ for forward propagation and backward propagation of single GPU; (2) $t_2$ for communication of weight gradients between GPUs or between GPU and CPU; (3) $t_3$ for preparation before training process and finishing work after training.

For Parameter Server framework, $t_1$ is inversely proportional to the number $n$ of GPUs used. $t_2$ is proportional to the number $n$ of GPUs and $t_3$ has nothing to do with $n$. So the training time metrics for PS framework $T_{PS}(n)$ is shown as below:
\begin{equation}
t=\frac{T}{n}+C \cdot n+P, \ n=2,3,\dots
\label{ps-time}
\end{equation}
Where $T$ is the training time with using single GPU, $C$ is the communication time and $P$ is the preparation time for opening and closing deep learning platform.

For our ring parallel architecture, $t_1$ is also inversely proportional to the number of $n$ of GPUs used. Let the size of $\Delta w$ in each GPU is $K$, single GPU send $\frac{K}{n}$ to his right neighbor each round. Every GPU do $n-1$ rounds of scatter and $n-1$ rounds of gather, so the total communication volume is $2K\cdot \frac{n-1}{n}$. Then we can get that $t_2$ is proportional to $\frac{n-1}{n}$. And $t_3$ is also a constant. So the training time metrics for ring parallel framework $T_{Ring}(n)$ is shown as below:
\begin{equation}
    t=\frac{T}{n}+C\cdot \frac{n}{n-1}+P, \ n=2,3,\dots
\label{ring-time}
\end{equation}

The most difference between these two training time metrics is $t_2$. As the number of GPUs $n$ increases, $t_2$ in ring parallel architecture will be smaller than PS framework.

\section{Experiments}
In this section, we evaluated our parallel temporal convolution network as two part. On the one hand, we tested the accuracy of proposal generation based on MTN; on the other hand, we evaluated the accelerated efficiency of our ring parallel architecture on MTN.
    
\subsection{Temporal action detection}
\textbf{Dataset. ActivityNet-1.3} \cite{caba2015activitynet} is a normal video dataset for the temporal action proposal generation task. It contains 19994 untrimmed videos containing 200 classes of action instance and corresponding annotations. each untrimmed video includes one or more action instances. ActivityNet-1.3 is divided into training set, test set and validation set in a ratio of approximately 2:1:1. \textbf{THUMOS14} \cite{jiang2014thumos} is a smaller video database containing 20 class action instances but they have a longer duration. THUMOS14 contains 213 and 200 temporal annotated untrimmed videos in testing and validation sets separately. In this part, we will compare the performance of various commonly used methods for temporal action detection on ActivityNet-1.3 and THUMOS14.

\textbf{Evaluation metrics.} Because it is rare that our generating proposals completely coincide with ground truth, we need to set a threshold of IoU (Intersection over Union) to judge whether proposals are correct or not. When the IoU between the proposal and ground truth is higher than the threshold, it is correct. In temporal action proposal generation task, Average Recall (AR) calculated with multiple IoU threshold is usually used as evaluation metrics. For ActivityNet-1.3, we set IoU threshold as [0.5: 0.05: 0.95] and [0.5: 0.05: 1.0] for THUMOS14. Because AR increases with the increase of AR with Average Number of proposals (AN), we use AR with definite AN as metrics, which is denoted as AR@AN. Besides, we also apply area under the AR vs. AN curve (AUC) as metrics On ActivityNet-1.3.

\textbf{Implementation details.} Two-stream network \cite{simonyan2014two} whose temporal network is BN-Inception \cite{ioffe2015batch} and spacial network is ResNet \cite{he2016deep} is used to encode videos. About the parameters in SENet\cite{hu2018squeeze}, we set reduction ratio with 512 feature map as 16 and reduction ratio with 3 feature map as 1. Besides, to test the impact of our improvement on final proposal generation result, we apply proposal generation module (PGM) and proposal evaluation module (PEM) in boundary sensitive network (BSN) to deal with the output of our temporal convolution network. The structure of BSN is shown in Fig. \ref{mtn} (c). We implement MTN with TensorFlow \cite{abadi2016tensorflow}. Our parallel computing platform are 8 TITAN V-100 GPUs.

\begin{figure}[ht]
    \centering
    \includegraphics[width=0.45\textwidth]{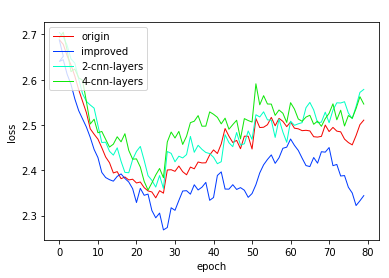}
    \caption{Loss curve with validation set. \textit{origin} is the original temporal convolution network and \textit{improved} denotes MTN. \textit{2-cnn-layers} and \textit{4-cnn-layers} mean to reduce or add one one-dimensional convolution layer based on \textit{origin}}
    \label{compare-tem-loss}
\end{figure}

We tested the loss curve with validation set between the original temporal convolution network and MTN. We also briefly changed the original architecture by adding or reducing one convolution layer The result is shown as Fig. \ref{compare-tem-loss}. We can see that whatever we increase or decrease the number of layers, the curve will be worse. But there is an obvious decline with using our improvement.

We also compared the final result of proposal generation between our MTN and other state-of-art proposal generation algorithms, which is shown as Table \ref{comapre-different-methods} and Table \ref{comapre-different-methods-thumos14}. 
\begin{table}[ht]
    \centering
    \caption{Comparison results between MTN and other state-of-the-art proposal generation methods on the validation set of ActivityNet-1.3 in terms of AR@AN and AUC.}
    \label{comapre-different-methods}
    \begin{tabular}{l*{2}c}
        \toprule
        Method & AR@100 (val) & AUC (val)  \\
        \midrule
        Zhao et al. \cite{zhao2017temporal} & 0.653 & 53.02 \\
        Dai et al. \cite{dai2017temporal} & - & 59.58\\
        Yao et al. \cite{ghanem2017activitynet} & - & 63.12\\
        Lin et al. \cite{lin2018bsn} & 0.748 & 66.17\\
        \textbf{MTN} & \textbf{0.756} & \textbf{67.26}\\
        \bottomrule
    \end{tabular}
\end{table}

\begin{table}[ht]
    \centering
    \caption{Comparison results between MTN and other state-of-the-art proposal generation methods on the validation set of THUMOS14 in terms of AR@AN. For simplicity, we use @AN instead of AR@AN.}
    \label{comapre-different-methods-thumos14}
    \begin{tabular}{l*{5}cr}
        \toprule
        Method & @50 & @100 & @200 & @500   \\
        \midrule
        DAPs \cite{escorcia2016daps} &   13.56 &  23.83  & 33.96 & 49.29     \\
        SCNN-prop \cite{shou2016temporal} & 17.22 & 26.17 & 37.01 & 51.57     \\
        SST \cite{buch2017sst} & 19.90 & 28.36 & 37.90 & 51.58   \\
        TURN \cite{gao2017turn} & 19.63 & 27.96 & 38.34 & 53.52  \\
        BSN \cite{lin2018bsn} & 29.58 & 37.38 & 45.55 & 54.67  \\
        MTN & \textbf{30.61} & \textbf{38.12} & \textbf{46.24} & \textbf{55.31} &  \\
        \bottomrule
    \end{tabular}
\end{table}

\textbf{Effectiveness of modules in MTN.} To evaluate the effectiveness of SENet and multipath dense layers, we demonstrate an ablation study on ActivityNet-1.3. The result is shown in Table \ref{ablation_study}. We can see from the table that the Multipath DenseNet has played a more effective role than SE-ConvNet in MTN, this is because the information extracted by Multipath DenseNet is a representation of information not just a promotion of existing information.
\begin{table}[ht]
    \centering
    \caption{Ablation study of MTN based on ActivityNet-1.3. MTN with SENet denotes the absence of dense layers and MTN with Multipath denotes abandoning SENet in MTN. For simplicity, we use @AN instead of AR@AN.}
    \label{ablation_study}
    \begin{tabular}{p{2.3cm}p{0.6cm}p{0.6cm}p{0.6cm}p{0.6cm}p{0.6cm}p{0.6cm}}
        \toprule
        Methods & @1 & @5 & @10 & @50 & @100 & AUC    \\
        \midrule
        Origin & 0.292 & 0.469 & 0.549 & 0.696 & 0.748 & 66.17    \\
        SE-ConvNet & 0.303 & 0.476 & 0.553 & 0.699 & 0.750 & 66.57    \\
        Mul-DenseNet & 0.317 & 0.482 & 0.556 & 0.702 & 0.751 & 66.85    \\
        MTN   & 0.332 & 0.490 & 0.562 & 0.706 & 0.756 & 67.26      \\
        \bottomrule
    \end{tabular}
\end{table}

\subsection{Parallel computing acceleration}
\begin{figure}[ht]
    \centering
    \includegraphics[width=0.45\textwidth]{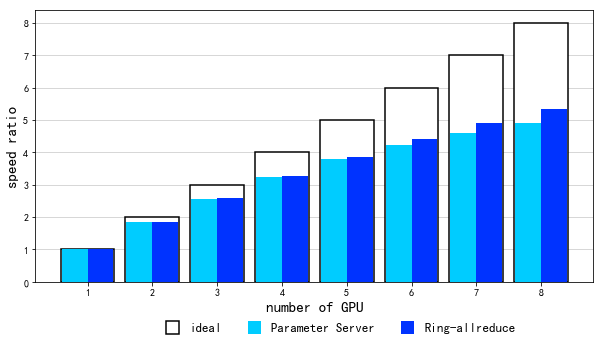}
    \caption{Comparison of speed ratio between PS and parallel ring architecture}
    \label{parallel_architecture_compare}
\end{figure}

Because of occupying huge memory and a large number of video files, distributed deep learning is indispensable for applying temporal action detection algorithm to the actual. Because of the inefficient of the traditional PS framework on distributed deep learning, we applied parallel ring architecture to our temporal convolution network and received a good result. The speed ratio with these two parallel frameworks is shown as Fig. \ref{parallel_architecture_compare}. As the number of GPU increases, the performance of parallel ring architecture is getting better and better than PS architecture. 

To further explore the relationship between the number of GPUs and training time, we defined training time function $T(n)$, where $n$ is the number of GPU used.

\begin{figure}[htpb] \centering    
\subfigure[Parameter Server architecture] {
 \label{ps-curve-fit}     
\includegraphics[width=0.8\columnwidth]{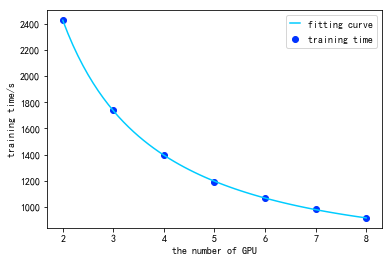}  
}     
\subfigure[Parallel ring architecture] { 
\label{ring-curve-fit}     
\includegraphics[width=0.8\columnwidth]{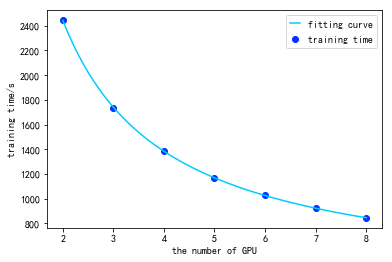}     
}     
\caption{Fitting curve of training time function}     
\label{curve_fit}     
\end{figure}

In order to test the validation of our training time metrics, we use the number of GPUs used, $n$, as the independent variable and the training time as the dependent variable to fit the Eq. \ref{ps-time}. The fitting curve is shown in Fig. \ref{ps-curve-fit} and we can get that $T=4223.8, C=12.1, P=290.8$.

If we use the same training time metrics \ref{ps-time} to fit the training time with using parallel ring architecture, we would get the result of $C=-3.8<0$, which is obviously unreasonable. The fitting curve for ring parallel architecture with using Eq. \ref{ring-time} is shown in Fig. \ref{ring-curve-fit} and we can get that $T=4400.1, C=59.6, P=363.5$.

The parameter $C$ is much bigger in parallel ring architecture than it in PS, which indicates that parallel ring architecture is lower than PS in transfer speed. But if the number of GPUs increases to a obvious level, especially in large scale deep learning like some video task, parallel ring architecture will be a better choice than Parameter Server framework.

\section{Conclusion}
In this paper, we proposed Multipath Temporal ConvNet (MTN) for proposal generation task and applied a new parallel architecture, ring parallel architecture, to accelerate our network by reducing the pressure of communication. Multipath Temporal Network can extract more effective information from long video feature sequences. Our state-of-the-art  method here doesn’t only have a better performance but have a higher acceleration efficiency compared with other action proposal generation methods, which is significant for dealing with large-scale video databases in industrial filed.

\Acknowledgements{This work is partially supported by the National Natural Science Foundation of China (61972016), the Fundamental Research Funds for the Central Universities (YWF-19-BJ-J237), the Open Research Fund of Fujian Engineering Research, Center of Public Service Big Data Mining and Application,
Fuzhou, China. The experimental platform is provided by Marc Casas at the Barcelona Supercomputing Center (BSC).}

\bibliographystyle{IEEEtran}
\bibliography{ref}

\end{document}